\documentclass[conference]{IEEEtran}
\IEEEoverridecommandlockouts
\usepackage[utf8]{inputenc}
\usepackage{cite}
\usepackage{amsmath,amssymb,amsfonts}
\usepackage{algorithmic}

\usepackage{graphicx}
\usepackage{subcaption}
\usepackage{siunitx}
\usepackage{textcomp}
\usepackage{xcolor}
\usepackage{enumitem}
\usepackage{url}
\usepackage{breakurl}
\usepackage[breaklinks]{hyperref}
\usepackage{tikz}

\usepackage{fancyhdr}
\usepackage[ruled,vlined]{algorithm2e}
\include{pythonlisting}
\def\BibTeX{{\rm B\kern-.05em{\sc i\kern-.025em b}\kern-.08em
    T\kern-.1667em\lower.7ex\hbox{E}\kern-.125emX}}
\begin{document}

\newcommand\copyrighttext{%
  \footnotesize \textcopyright 2021 IEEE. Personal use of this material is permitted.
  Permission from IEEE must be obtained for all other uses, in any current or future 
  media, including reprinting/republishing this material for advertising or promotional 
  purposes, creating new collective works, for resale or redistribution to servers or 
  lists, or reuse of any copyrighted component of this work in other works. 

Accepted to be published in: 2021 International Joint Conference on Neural Networks (IJCNN 2021), July 18-22, 2021.
}
\newcommand\copyrightnotice{%
\begin{tikzpicture}[remember picture,overlay]
\node[anchor=south,yshift=10pt] at (current page.south) {\fbox{\parbox{\dimexpr\textwidth-\fboxsep-\fboxrule\relax}{\copyrighttext}}};
\end{tikzpicture}%
}

\title{A study on Ensemble Learning for Time Series Forecasting and the need for Meta-Learning}

\author{
    \IEEEauthorblockN{Julia Gastinger, S\'{e}bastien Nicolas, Du\v{s}ica Stepi\'{c}, Mischa Schmidt, Anett Sch\"ulke }
    \IEEEauthorblockA{NEC Laboratories Europe GmbH, Kurf\"ursten-Anlage 36, 69115 Heidelberg, Germany
    \\\{FirstName.LastName\}@neclab.eu}    

}

\maketitle
\copyrightnotice

\begin{abstract}
The contribution of this work is twofold: (1) We introduce a collection of ensemble methods for time series forecasting to combine predictions from base models. We demonstrate insights on the power of ensemble learning for forecasting, showing experiment results on about 16000 openly available datasets, from M4, M5, M3 competitions, as well as FRED (Federal Reserve Economic Data) datasets. Whereas experiments show that ensembles provide a benefit on forecasting results, there is no clear winning ensemble strategy (plus hyperparameter configuration). Thus, in addition, (2), we propose a meta-learning step to choose, for each dataset, the most appropriate ensemble method and their hyperparameter configuration to run based on dataset meta-features. 
\end{abstract}


\section{Introduction}\label{sec:intro}
Anything that is observed sequentially over time is a time series \cite{hyndman2018book}. 
When forecasting time series data, the aim is to estimate how the sequence of observations will continue into the future\cite{hyndman2018book}. The datasets typically consist of a timestamp, a target value, plus - possibly- exogenous variables. In the presented work we only consider time series that are observed at regular intervals of time.

The motivation leading to this work is the creation of a platform for automated time series forecasting. 
This platform should leverage forecasting algorithms and combine their predictions with promising ensemble method(s) to provide a forecast for any given time series, possibly including exogenous variables. 

The literature agrees on the value of ensemble learning, see e.g. \cite{montero2020}, \cite{makridakis2018m4}, \cite{makridakis2020forecasting}, \cite{timmermann2006forecast}. 
Various promising ensemble learning methods for forecasting exist (see section \ref{lit:ensembles}). 
Anyway, we could not find a recent study that collects and thoroughly compares various existing ensemble learning methods for forecasting different datasets on various use cases.
For this reason, we collected and implemented a diverse assortment of ensemble methods suited for time series forecasting with and without exogenous variables.
To compare their performance, and eventually decide which (subset of) ensemble methods to include in the platform, we run experiments on a large and diverse amount of time series. 
The collected methods and experiment results constitute the first part of this paper.

We can confirm that ensemble learning benefits forecasting accuracy. However, we find that  
preselecting one - or a subset of - ensemble method(s) for time series forecasting tasks in order to reach high accuracy (while avoiding running all instantiations of ensemble methods in parallel) is a question of horses for courses.
Thus, in the second part, we propose to use a meta-learning approach:
 It should preselect, for each (new) time series, one or multiple ensemble methods plus hyperparameter setting, based on the dataset's meta-features. 

This work contributes:
(1) a collection of ensemble learning methods to combine predictions from forecasting models and to integrate exogenous variables,
(2) an experimental evaluation on a large and diverse amount of (openly available) datasets, 
(3) a discussion of presented approaches, and,
(4) a meta-learning step to select the expected best subset of ensemble learning methods plus hyperparameters for each new dataset, based on dataset meta-features. 

\section{Related Work}\label{sec:related}
\subsection{Time Series Forecasting and Forecasting Competitions}\label{sec:mcomp}
For an overview of time series forecasting, we refer to \cite{hyndman2018book}. 
The most famous competitions organized to improve the forecasting accuracy and make more significant contributions 
in the field of forecasting are the M competitions (e.g. \cite{makridakis2000m3}, \cite{makridakis2018m4}, \cite{makridakis2020m5}). The M competitions are interesting for this work due to the provided datasets, the findings, and the presented methods.
\subsubsection{M3 competition}
The M3 competition~\cite{makridakis2000m3} (1999) totals 3003 datasets (yearly, quarterly, monthly, other) in various categories. 
The winning method was the theta method, a purely statistical exponential-smoothing-based algorithm\cite{Assimakopoulos2000}. 
\subsubsection{M4 competition} 
The M4 competition~\cite{makridakis2018m4} (2018) bases on 100000 real-life time series (yearly, quarterly, monthly, daily, hourly) from various categories. 
\cite{makridakis2018m4} define the major contributions of M4 as: (a) The introduction of a “hybrid” (statistical + neural network) method; (b) the confirmation of the superiority of ensemble methods (referred to as combinations) and the incorporation of ML algorithms for determining their weighting, which improves their accuracy; and (c) the impressive performances of some methods for determining prediction intervals. In conclusion, \cite{makridakis2018m4} define the utilization of hybrid and ensemble methods as the logical way forward.
\subsubsection{M5 competition}
The M5 competition~\cite{makridakis2020m5} (2020) focused on a retail sales forecasting application. It bases on hierarchical sales data, totaling 42840 series, including exogenous/explanatory variables (including calendar-related information, selling prices, promotion activities) in addition to the time series data, using grouped, correlated time series. The daily data covers a period from 2011 to 2016. 
Results from this competition show: (a) The superiority of simple ML methods on such a dataset; (b) the value of ensemble methods; (c) the value of cross-learning; (d) the beneficial effect of external adjustments; (e) the value added by effective cross-validation; and (f) the importance of exogenous/explanatory variables.
\subsubsection{Relation to this work}
In the presented work we use datasets from M3, M4, and M5. We use datasets from multiple competitions to test our approach with multiple seasonalities and for multiple use cases, for time series with and without exogenous variables.
We include the theta model \cite{Assimakopoulos2000}, the winner of the M3 competition, as a base model.
We do not work on hybrid methods in this presented work but instead focus on ensemble methods, which were identified as promising in the findings of the M4 competition.
Thus, we do not integrate the first winner of the M4 competition, a hybrid method, but integrate the runner-up, \cite{montero2020}, as one of the ensemble methods as well as ideas of the third winner \cite{pawlikowski2020}. 
Contrary to the M5 competition, our work does not focus on taking into account the hierarchical structure and grouping/correlation of time series. Neither does this particular investigation use machine learning models as base models. Thus, we do not focus on the methods presented in the M5 competition. We integrate machine learning models in some ensemble strategies and in the meta-learning step.
In addition to including multiple findings from the M competitions, we want to provide an approach that is not tailored to one specific competition but instead, provide an all-rounder approach. 

\subsection{Ensemble Methods for Time Series Forecasting}\label{lit:ensembles}
For a collection of ensemble methods used in this paper see section \ref{lit:ensembles}. 
A major challenge for ensemble learning lies in selecting an appropriate set of weights, and many attempts to do this have been worse than simply using equal weights \cite{montero2020}. 
Further, the selection of base models for the ensemble from a pool of base models is considered a challenging task, tackled e.g. by \cite{pawlikowski2020}, \cite{caruana2004}. 
Whereas there are many methods for ensemble learning in time series forecasting, to the best of our knowledge, no study exists which collects and thoroughly compares various ensemble methods for forecasting on different datasets.

\subsection{Meta-learning for Time Series Forecasting}
As one of the first, \cite{michie1994}, use meta-learning for algorithm selection in machine learning. 
For each algorithm to be selected they created one meta-learning binary classification problem, where the class labels were assigned by comparing results to a confidence interval. 
There have been several attempts to use time series features in combination with meta-learning for forecasting:
\cite{prudencio2004} apply meta-learning to select models for time series forecasting, investigating two approaches: a single machine learning algorithm to select among two models to forecast stationary time series, and the NOEMON approach \cite{kalousis1999noemon}, to rank three models used to forecast time series of the M3-Competition. \cite{graff2013} also propose two ways of selecting a forecasting algorithm: selection by predicting each forecaster's performance with a linear combination of a) features that were previously used to assess the problem difficulty of evolutionary algorithms, together with b) an additional set of time series features; and selection by treating the process as classification task, using the proposed features to describe each time series.
The approach in \cite{lemke2010} aims, in addition to selecting a model, also at deciding whether to pick an individual model or a combination. 
More recently, \cite{talagala2018} introduced a framework that uses time series features combined with meta-learning for forecast-model selection. 
Building on \cite{talagala2018}, \cite{montero2020} present FFORMA (Feature-based FORecast Model Averaging), the runner-up of the M4 competition, which uses meta-learning to select the weights for a weighted forecast ensemble. 

To the best of our knowledge, there is no work that uses meta-learning to select ensemble methods plus their hyperparameter configuration for time series forecasting.

\section{Ensemble Learning for Forecasting}\label{sec:methods}
\subsection{Time Series Forecasting Platform}
The objective of our platform is to automatically find the \textit{best} algorithm to train for 
each time series dataset among a pool of available algorithms, including ensemble methods and 
their hyperparameter configuration. We define the \textit{best} algorithm to be the one 
yielding the lowest error on the test set. 
The user can upload a time series dataset and define a train-test split. The platform 
automatically runs a preprocessing step (section \ref{sec:preprocessing}), fits the base 
models (section \ref{sec:basealgos}), and runs ensemble methods (section \ref{sec:ensemble}). 
It returns the model with the lowest test error. Whereas it is possible to run all base models 
and ensemble methods, to find the highest-scoring model, we prefer to preselect a 
sub-set of (promising) methods to save CPU power and memory. This can either be done 
manually by the user or by an integrated meta-learner (section \ref{sec:meta}). 
\subsubsection{Preprocessing}\label{sec:preprocessing}
We implement two Preprocessors. 

\textit{Seasonality Detection:} First, the preprocessor infers seasonality by computing a periodogram 
(fast Fourier transform) of the target variable and selecting the first spectral density peak.
It then removes this first seasonality from the dataset (a check is carried out to select 
between additive and multiplicative seasonality) and carries out the same process to identify 
a potential second seasonality. 
Second, if the (first or second) seasonality is more than a certain threshold, it creates
Fourier terms for the automated ARIMA algorithm instead of letting the algorithm run with the 
actual seasonality (i.e. a high number of linear terms to optimize over).

\textit{Exogenous Variables:} The preprocessor extracts numerical/categorical 
variables from the date/time column if it is present, for example, time of the day, day of the week, 
etc. It also extracts holidays and workday boolean variables, if information about the country is 
provided.

\subsubsection{Base Algorithms}\label{sec:basealgos}
Following the work of \cite{montero2020}, we consider the base algorithms in Table \ref{tab:base}. For all methods, we use the default settings. The preprocessor described in section \ref{sec:preprocessing} provides seasonality information.
%
\begin{table}[]
\scriptsize
\caption{Base Algorithms, Packages Used and Sources.}
\label{tab:base}
\begin{tabular}{>{\raggedright}p{5.3cm} >{\raggedright}p{0.7cm} p{1.5cm} }  
Algorithm & Package & Reference \\\hline 
Automated ARIMA (arima) & \cite{pmdarima2017} & \cite{hyndman2018book}, sec 8.9 \\
Theta method (theta) & \cite{loning2020}& \cite{Assimakopoulos2000} \\
TBATS models (tbats) & \cite{tbats2019} & \cite{de2011} \\
Automated exponential smoothing algo. (ets) & \cite{loning2020} & \cite{hyndman2018book}, sec 7.7\\
Neural network time series forecasts (R\_nnetar) & \cite{hyndman2018} & \cite{hyndman2018book}, sec 11.3 \\
Seasonal and Trend decomposition using Loss with AR modeling of the seasonally adjusted series (R\_stlmar) & \cite{hyndman2018}& \cite{hyndman2018book}, sec 6.6\\
Random walk with drift (R\_rwdrift) &\cite{hyndman2018}  & \cite{hyndman2018book}, sec 3.1\\
Na\"ive (naive) & Own & \cite{hyndman2018book}, sec 3.1 \\
Seasonal Na\"ive (snaive) & Own  & \cite{hyndman2018book}, sec 3.1\\
\end{tabular}
\end{table}

\subsection{Ensemble Methods}\label{sec:ensemble}
For combining predictions from base models, as well as enhancing the predictions with information from exogenous variables, we implement a variety of ensemble methods. 
They vary in their model selection strategy, in their way of computing weights, in their meta-model, and in their way of integrating exogenous variables.
To select which and how many base models to be combined by the ensembles, we implement different selection strategies: all, best, algo\_name. The strategy all selects all base models from the models pool, best selects the 
$num\_best$ models with the lowest training error ($num\_best$ is a configuration parameter, set to $b=3$), and algo\_name selects the base algorithm with the specified name. Please note that this step precedes running the ensemble methods and should not be confused with the ensemble methods specialized in model selection as described in sections \ref{ens:modelsel}, \ref{ens:bags}, \ref{ens:backwardel}. Table \ref{tab:base} shows the combinations of ensemble methods and selection strategies that we use in this work. We identified them as most promising in preceding experiments (not contained in this work).

Each ensemble method comes with hyperparameters. After running a set of preceding experiments (not contained in this work) we were able to exclude non-promising hyperparameter configurations. Because we were not able to distinguish clear winning configurations, we keep multiple configurations for each method. See table \ref{tab:hyper} for an overview of each ensemble method with its open hyperparameters and possible values. 
For ensemble methods that need an additional data split (for training and validation) we split the original training data $\boldsymbol{X}_{\textrm{train}}, \boldsymbol{y}_{\textrm{train}}$ in an ensemble train set $\boldsymbol{X}_{\textrm{etrain}}, \boldsymbol{y}_{\textrm{etrain}}$ and ensemble validation set $\boldsymbol{X}_{\textrm{evalid}}, \boldsymbol{y}_{\textrm{evalid}}$.

\begin{table}[]
\scriptsize
\caption{Ensemble Methods, associated Hyperparameters and Selection Strategies.}
\label{tab:hyper}
\begin{tabular}{>{\raggedright}p{2.65cm} >{\raggedright}p{3.5cm} p{1.3cm} } 
Ensemble Method												& Hyperparameters \{Values\}  														& 	Selection \newline Strategies \\\hline 
combine\_detwe												& formula=\{sqr, inv, exp\}																& \{all, best\} 	\\ 
ensemble\_stacking											& meta-model=\{linreg, rf, lgbm, xgboost\}\  										& \{all, best\} 	\\   
ensemble\_stacking\_basic	  								& meta-model=\{linreg, rf, \newline lgbm, xgboost\} 								& \{all, best\} 	\\   
ensemble\_model\_selection 								& sort=\{true, false \}																	& \{all, best\} 	\\   
ensemble\_selection\_bags	  								& metric =\{mean\_squared\_error, mean\_absolute\_error \}					& \{all, best\} 	\\   
ensemble\_backward\_ \newline elimination				& combination=\{weighted\_average, stacking\}, \newline stacking-meta-model=\{linreg, rf, lgbm, xgboost\} & \{all\} \\ 
fforma 															& - 																							& \{all\}\\ 
recent\_ensemble 												& P=\{20, 50\}, $\lambda$=\{20, 30, 50\}  										& \{all\} \\ 
superbooster 													& noise=\{true, false\}, \newline  meta-model=\{lgbm, xgboost\}  			& \{tbats, best, naive\}\\  
mean\_average 												& -																								& \{all, best\}\\  
algo\_algo 														& -																								& \{ets\_arima, ets\_arima\_\newline tbats\_theta, ets\_arima\_tbats\}\\ 
\end{tabular}
\end{table}

\subsubsection{Compute Weights based on performance (combine\_detwe)}\label{ens:detwe}
The method determines the combination weights based on the base models validation score (sMAPE error), using one out of three provided formulas introduced in \cite{pawlikowski2020}. Which formula to use, is a hyperparameter in our case.

\subsubsection{Train model to combine base model predictions (ensemble\_stacking)}\label{ens:stack}
This ensemble technique is based on the application of stacked generalization to k-fold cross-validation, the Super Learner \cite{van2007super}, adapted to time series with a time series cross-validator. It trains a machine learning model to learn how to combine the predictions of the base models. For that it evaluates each base model using the time series cross-validation, stores the out-of-fold predictions, and fits the base model on the full training dataset. It then fits the meta-model on the out-of-fold predictions. Which model to use as meta-model is in our case a hyperparameter. We choose between linear regression, random forest, LGBM, and XGBoost, where we use the default configuration for each model.

In addition, as a less computing-intensive alternative ensemble technique better suited to time series with small sample sizes, we train a meta-model, but instead of using the time series cross-validator we use simple train test subset splits without shuffling (ensemble\_stacking\_basic). This is less computing-intensive and should work better with time series with a small number of samples.

\subsubsection{Ensemble Forward Selection (ensemble\_model\_selection)}\label{ens:modelsel}
This idea, originally introduced by \cite{caruana2004}, uses forward stepwise selection to add to the ensemble the models from the models pool that maximize ensemble-performance. The technique starts with the best model from the library (in our case the model with the lowest validation error). In each iteration, it adds another model if its addition would improve the validation score. Repeated addition of a model is possible. We compute the weight of the model based on the number of times it was added.

\subsubsection{Ensemble Selection Built on Bags (ensemble\_selection\_bags)}\label{ens:bags}
This method is similar to \ref{ens:modelsel}, with an adaption inspired by, but not equal to, the idea introduced in \cite{khiari2018metabags}. Instead of selecting and assigning weights based on the performance on the whole dataset as in \ref{ens:modelsel}, this idea only considers subsets of the entire dataset (bags).
Instead of creating the bags randomly, it builds them based on seasonality information, using the value from the preprocessor \ref{sec:preprocessing}: For e.g. a weekly series, the method creates seven bags. It applies the method \ref{ens:modelsel} for each bag. The method bases on the underlying assumption that a model which performed well at e.g. the preceding Sundays might also perform well on this Sunday.

\subsubsection{Ensemble Backward Elimination (ensemble\_backward\_elimination)}\label{ens:backwardel}
This method introduced by \cite{pawlikowski2020} is an alternative technique to select models from the base models pool. Instead of starting with only one model, as described in \ref{ens:modelsel}, it starts with a combination of all base model predictions. It removes the methods with the highest error on the validation set one by one, if, and as long as, that decreases the error. The remaining methods then constitute the ensemble base models, combined either using equal weights weighted average or using an additional stacking step as described in \ref{ens:stack}. The combination technique is in this case a hyperparameter. 

\subsubsection{Ensemble Meta-Learning (fforma)}\label{ens:fforma}
This method, FFORMA, introduced by \cite{montero2020}, is an automated method for obtaining weighted forecast combinations using time series features. The meta-learning model learns to produce weights for all methods in the pool, as a function of the features of the time series to be forecast, by minimizing summary forecast loss measure. This method was the runner-up of the M4 competition. 

\subsubsection{Ensemble based on performance of most recent data points (recent\_ensemble)}\label{ens:recent}
This technique, adapted from \cite{cerqueira2017}, selects and combines base models based on the performance on the most recent data points from the training dataset. The technique selects the best-performing $\lambda \%$ models and computes their weights by taking into account their performance on the $P$ most recent points of the train dataset. The parameters $\lambda$ and $P$ are therefore hyperparameters. Contrary to \cite{cerqueira2017}, we use the sAPE (sMAPE without mean) instead of the Squared Error to compute the performance score for scaling reasons, and do not update our train set over time. 

\subsubsection{Exogenous Variable Postprocessor (superbooster)}\label{ens:super}
This method\footnote{Technically this is not an ensemble method, as it only builds on one base model, but a postprocessor. For consistency, we list it with ensemble methods.} is a machine learning model, trained on top of a base model. It includes exogenous variables. The training dataset for the model consists of the base algorithm's predictions on the validation set, extended by the respective exogenous variables for each time step. The exogenous variables consist of the variables extracted by the preprocessor from section \ref{sec:preprocessing} plus, possibly, the dataset's original exogenous variables if existent. The target variables are the original target variables.

In addition, to create more training data, we implement a version that creates artificial additional training targets $\boldsymbol{y}_{\textrm{noisy}}$ for the meta-learner (set by the hyperparameter noise=True). 
$
\boldsymbol{y}_{\textrm{noisy}} = \boldsymbol{y}_{\textrm{etrain}}  + \alpha \: \boldsymbol{r}_{\textrm{v}},
$
with
$\boldsymbol{r}_{\textrm{v}} {\sim} \mathcal{N}(\boldsymbol{0},\,\boldsymbol{\Delta y})\in\mathbb{R}^{L\times1}$, 
and $\boldsymbol{y}_{\textrm{noisy}},\boldsymbol{y}_{\textrm{etrain}}\in\mathbb{R}^{L\times1}$, where $L$ is the number of training samples, and $\alpha$ a scaling factor (we set $\alpha=0.1$); Further,
$
\boldsymbol{\Delta y} =\mathbf{1} \frac{1}{V} \sum_{i=1}^{V}{|\hat{\boldsymbol{y}}_{\textrm{evalid},i} - \boldsymbol{y}_{\textrm{evalid},i}|}, 
$
with, $\hat{\boldsymbol{y}}_{\textrm{evalid}}, \boldsymbol{y}_{\textrm{evalid}} \in \mathbb{R}^{V\times1}$,  $V$ the number of validation samples,  $\boldsymbol{y}$ the actual target values, and $\hat{\boldsymbol{y}}$ the target values as predicted by the base model.
The new artificially extended training target vector for training the meta-learner is constructed as follows: $\boldsymbol{y}_{\textrm{extended}} = [\boldsymbol{y}_{\textrm{noisy}}; \boldsymbol{y}_{\textrm{evalid}}]^{\textrm{T}}$, with $\boldsymbol{y}_{\textrm{extended}} \in \mathbb{R}^{(L+V)\times1}$.

\subsubsection{Mean Average (mean\_average)}\label{ens:mean}
This computes the mean across all base models' predictions with equal weights.

\subsubsection{Fixed Combinations (algo\_algo)}\label{ens:fixed}
This combines the predictions of specified time series algorithms with equal weights. Due to the most promising results in preceding experiments we use the combinations as in table \ref{tab:hyper}.

\subsection{Validation - Experiment 1}\label{sec:valensemble}
\subsubsection{Research Questions}
\begin{enumerate}[label=(\roman*)]
\item Do ensemble methods help to improve the forecasting accuracy?
\item With the goal of reaching the best possible forecasting accuracy for automated domain-agnostic time-series forecasting, and the possibility to run multiple ensemble methods in parallel: Which ensemble methods (including hyperparameter configurations) should be selected? 
\item Do the results differ significantly for different dataset-sources (the different M competitions, FRED)?
\end{enumerate}

\subsubsection{Scope, Methodology}
In the first set of experiments we aim to compare the forecasting performance of different ensemble+HPs, and base models from a domain-agnostic perspective. The term ensemble+HP describes in this case an ensemble method and hyperparameter configuration.  
We run the base models on each of the datasets described in \ref{sec:datasets} and combine their predictions running each of the ensemble+HPs. 

We use a customized version of the symmetric Mean Absolute Percentage Error (sMAPE) metric. As the standard sMAPE is ill-suited when dealing with a large fraction of zeros as target values (as is the case for the M5 datasets) we modify it to better handle this problem and to be better protected against outliers compared to the version introduced by \cite{armstrong1985}. The customized version is defined as:
\begin{equation}
\text{sMAPE} = 100 * \frac{\sum_{t=1}^{k}{|\hat{y}_t-y_t|}}{|\sum_{t=1}^{k}{(\hat{y}_t +y_t)}|},
\end{equation}
with $k$ being the number of test samples, $y$ the actual value, and $\hat{y}$ the forecast value. 
Because we want to find out which ensemble+HPs to select for a future platform, we compare the methods looking at their mean ranks, as well as the number of wins. The advantage of comparing methods by their ranks and number of wins over comparing them by their sMAPE is the better comparability across datasets.
In our experiments, we do not consider training or execution time.

\subsubsection{Datasets, Computational Resources, and Setup}\label{sec:datasets}
We run the base learners described in section \ref{sec:basealgos} and ensemble+HPs as described in section \ref{sec:ensemble} on the following datasets\footnote{Because of limited available CPU time we were only able to run a subset of all available datasets. List of datasets available on e-mail request.}:
A random subset of M3 datasets, \cite{makridakis2000m3}, containing 2039 time series, of M4 datasets, \cite{makridakis2018m4}, containing 9668 time series, and of M5 dataset (produce-level), \cite{makridakis2020m5}, containing 2997 time series. In addition, we add 1767 FRED (Federal Reserve Economic Data) datasets\cite{fred2020}. Please note that it is out of scope of our work to take the hierarchical structure and correlation of time series as in the M5 competition into account. Instead, we look at each extracted series separately.

We ran parallel trainings of the ${\sim}16000$ datasets over a period of 3 weeks using 400 CPU cores from 25 physical Intel machines. We limited each base model or ensemble training to 1 CPU core and 2GB of memory, and did not limit training time of individual algorithms.

\subsection{Results and Discussion - Experiment 1}\label{sec:experiments1}
\subsubsection{Results}
\begin{figure*}[!h]
 \centering
\includegraphics[width=1.0\textwidth]{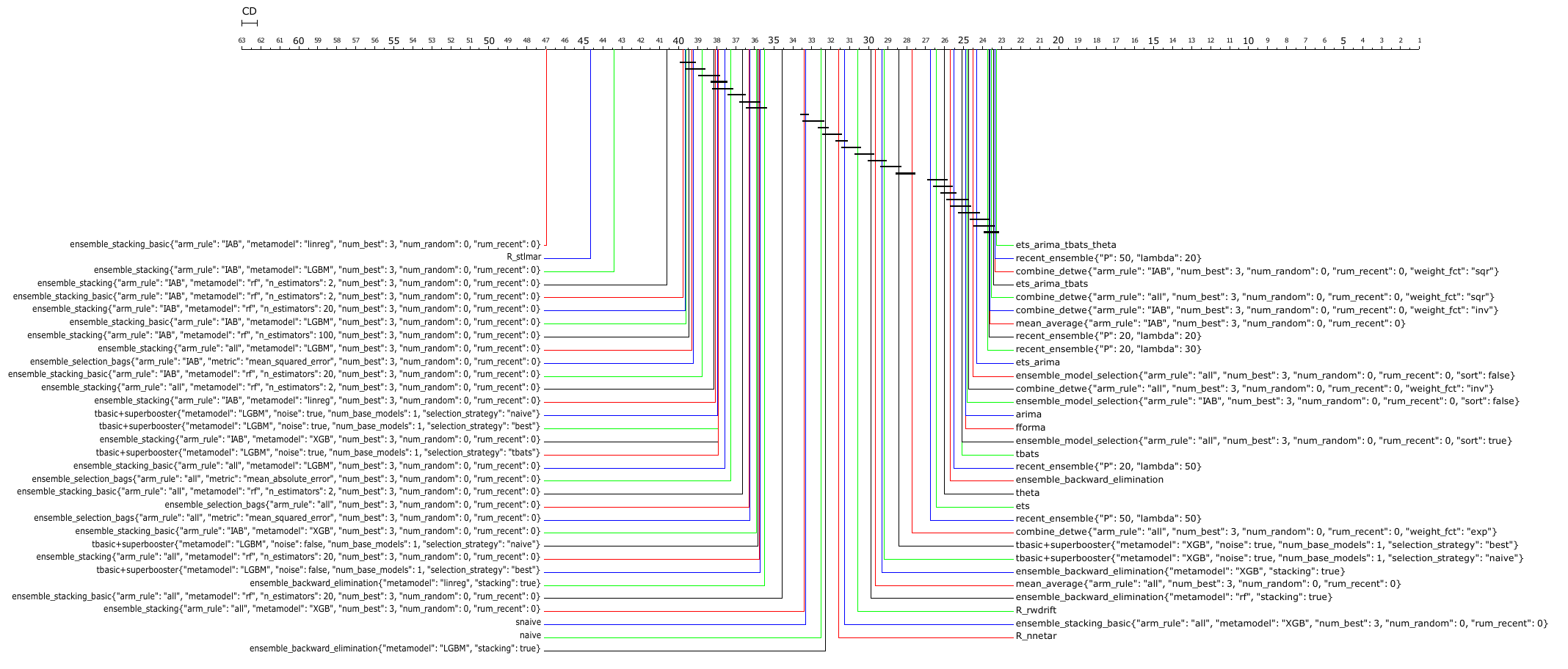}
\caption{Statistical significance diagram - best viewed zoomed, full size available by email request. It presents average ranks of algorithms (including hyperparameter configurations) across all datasets. Bold lines indicate groups of algorithms which are not significantly different, their average ranks differ by less than critical difference (CD) value. Colors for better readability.}
\label{fig:totalCI}
\end{figure*}

\begin{figure*}
 \centering
\includegraphics[width=1\textwidth]{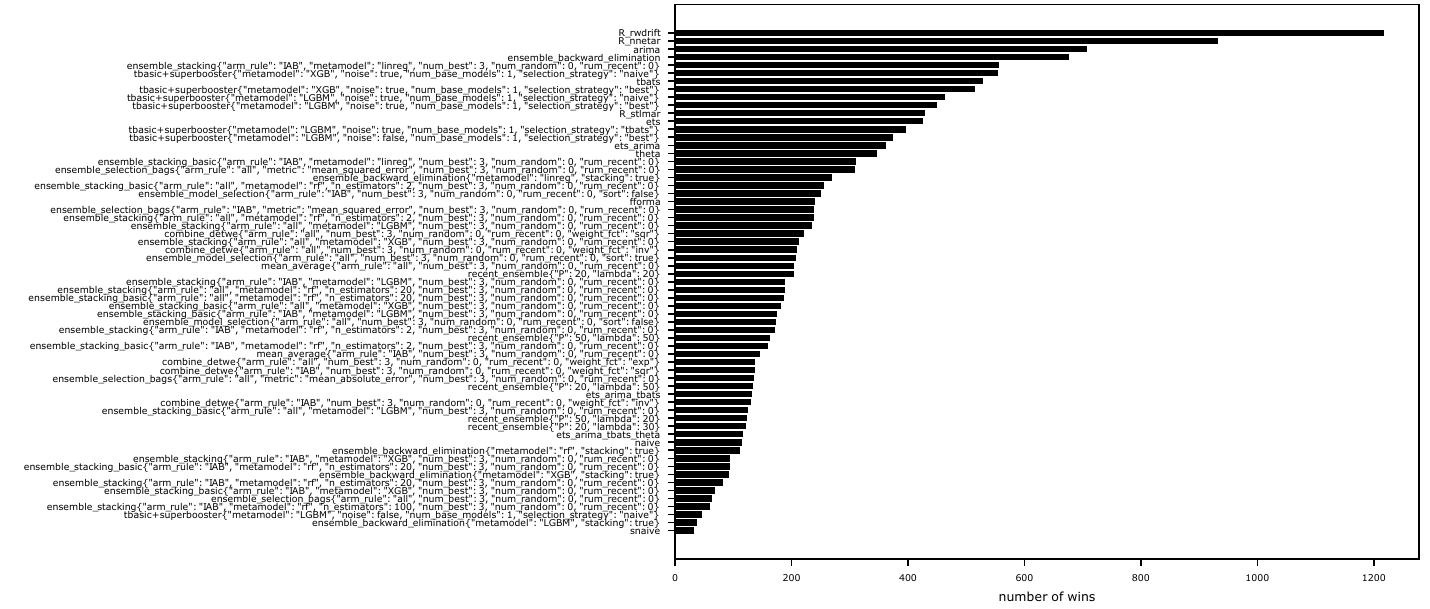}
\caption{Number of wins per algorithm across all datasets - best viewed zoomed, full size available by email request.}
\label{fig:nrwins}
\end{figure*}

For results describing the average rank of algorithms across all datasets please see the statistical significance diagram \cite{demvsar2006} in figure \ref{fig:totalCI} that shows statistically significant distance between groups of algorithms. The group with the lowest ranks contains only ensemble+HPs, no base models. 
The group only contains ensemble methods that compute their weights based on weighted average, with either equal weights or weights based on formulas, no machine learning model is trained to combine the models. 
The best algorithm has a rank of ${\sim}23$, the worst algorithm has a rank of ${\sim}47$. 
Please note that in our experiments we evaluate based on the customized sMAPE (in contrary to M4 experiments that evaluate based on OWA). 
For results describing the number of wins per algorithm across all datasets please see figure \ref{fig:nrwins}. The three algorithms with the highest number of wins are base algorithms. 
In 4735 out of 16471 times ($28.7\%$), the winning method is a base algorithm, in 11736 times ($71.3\%$) an ensemble method.

In addition, we evaluate the performance on subsets of the dataset, grouped by data-sources (M3, M4, M5, FRED) based on their ranks\footnote{For space reasons, we omit the figures describing the results and highlight only the most important findings in this work. Figures are available on request.}. 
For M3 we find that theta, the original M3-winner is with a mean rank of ${\sim}26$ in the group with lowest ranks according to the CD-evaluation. 
Also, fforma (Ensemble Meta-Learning)  is in the group of lowest ranks for M3 (mean rank of ${\sim}24$, second lowest rank) and M4 data (mean rank of ${\sim}23$, 14th lowest rank). 
The sub results on M5 show that three instantiations (different hyperparameter configurations) of superbooster (Exogenous Variable Postprocessor), which incorporates exogenous variables,  have the three lowest mean ranks, all with mean ranks of ${\sim}26$. This also includes superbooster who builds on the base model naive. For M5, fforma (Ensemble Meta-Learning) reaches a mean rank of ${\sim}35$, the 50th lowest rank - in contrary to e.g. the simple ets\_arima\_tbats combination (Fixed Combination) which is in the lowest-ranked group with a mean rank of ${\sim}27$, and to the base models arima (mean rank ${\sim}29$), theta (mean rank ${\sim}29$), or ets (mean rank ${\sim}30$). We observe across data-sources that the simple ensemble+HPs, eg. mean\_average, recent\_ensemble (Ensemble based on performance of most recent data points), combine\_detwe (Compute weights based on performance) reach mean ranks in the lower third of achieved ranks.

\subsubsection{Discussion and Findings}
In the following, we reply to the research questions in section \ref{sec:valensemble}.

(i): Due to the results of Experiment 1 we can confirm that ensemble methods help to improve the forecasting accuracy. The group of algorithms with the lowest rank only contains ensemble methods. In addition, in $71.3\%$ of time series, the best model was found by an ensemble model.

(ii): Whereas we can see a statistically significant distance in the mean ranks, it is not clear which ensemble methods to select. The best model's mean rank is ${\sim}23$, which we consider as suboptimal. 
In addition, algorithms with low mean rank do not necessarily have many wins. E.g. ets\_arima\_tbats\_theta (Fixed Combination), the algorithm with the lowest mean rank, is in the lower fourth in the category of number of wins.
Thus we conclude that models that do forecast with an \textit{in average low} sMAPE, are not necessarily the models with \textit{the lowest} sMAPE on many datasets. 
If the user wants to select a safe option, scoring well on average, we recommend choosing a simple ensemble technique, e.g. mean\_average, recent\_ensemble (Ensemble based on performance of most recent data points), fixed algo\_algo combinations or combine\_detwe (Compute weights based on performance), with hyperparameter configurations as in figure \ref{fig:totalCI}. This is not the case if the user wants to find \textit{the best} option for each dataset. Which ensemble method performs best varies across datasets. We cannot define a simple heuristic which ensemble method to run for which dataset by simple observation of the results. Thus, from the results in Experiment 1 we are not able to give a clear recommendation on which ensemble+HPs to select -  if resources were unlimited, it would lead to the best results to run all ensemble+HPs.

(iii): Due to the results of Experiment 1 we can confirm that the results differ for different data-sources. Where the algorithms which performed well in previous forecasting competitions, e.g. fforma (Ensemble Meta-Learning) do show low mean ranks in M3, M4 and FRED datasets, they do show high mean ranks for M5, even compared to some base algorithms. For fforma we suspect the reason to be that it was trained on M4 data, and that the fforma-meta-features do not capture the M5 time series with exogenous variables well. For the M5 datasets, which include exogenous variables, the superbooster (Exogenous Variable Postprocessor), including superbooster on top of base model naive reaches low ranks. This is in line with the findings of M5 competition, that Machine Learning algorithms help on those datasets. We find that - data-source independent - simple ensemble methods, e.g. mean\_average, recent\_ensemble (Ensemble based on performance of most recent data points), fixed algo\_algo combinations or combine\_detwe (Compute weights based on performance) show low average ranks.

\section{Meta-Learning for Ensemble Selection}\label{sec:methods}
\subsection{Method for Meta-learning}\label{sec:meta}
Due to our findings from Experiment 1, we propose a meta-learning step to select the ensemble+HPs to be used for a given dataset based on the dataset meta-features. Following the idea presented by \cite{michie1994}, we fit one classifier 
per ensemble+HP. 
In total, we apply 54 classifiers, each fitted on 14690 time series forecasting problems. 
This means that for each ensemble+HP we create one meta-learning classification problem. In this work, we only examine the situation where all base models are preselected and where the meta-learner selects a promising subset of ensemble+HPs.

\subsubsection{Dataset for Meta-learning}\label{sec:metadata}
To create the dataset for meta-learning we use the ensemble+HP performance from Experiment 1. 
Each meta-learning-dataset sample describes one time series. The features of the meta-learning dataset are the meta-features of the respective time series. We extract the 42 meta-features proposed in \cite{montero2020} with the package \cite{tsfeatures2020}. Because those meta-features were introduced for univariate time series without exogenous variables, we extend them with 7 common non-time-series-specific meta-features to take into account information about the existence and properties of exogenous variables. We use a subset\footnote{We only use a subset to avoid overlap with the time series meta-features, which would e.g. be the case for \textit{nr\_inst} (number of instances in the dataset).}
 of the general dataset meta-features from \cite{alco2020}: \textit{nr\_cat} (number of categorical attributes), \textit{nr\_bin} (number of binary attributes), \textit{nr\_num} (number of numeric features), \textit{nr\_attr} (total number of attributes), \textit{inst\_to\_attr} (ratio between the number of instances and attributes.),  \textit{num\_to\_cat} (number of numerical and categorical features).
We create one target column per ensemble+HP $y(ensemble+HP)$. When training the ensemble+HPs' meta-learner the respective target column is used. The target value is binary, being $1$ (select), or $0$ (do not select). We rank the ensemble+HPs based on their sMAPE and define the target $y$ as follows:
\[
    y(ensemble+HP)= 
\begin{cases}
    1,& \text{if } rank(ensemble\text{+}HP)\leq K\\
    0,              & \text{otherwise},
\end{cases}
\]
with configuration parameter $K$ depending on the user's preferences and availability of CPUs/memory.
We split the dataset to train/test set.
\subsubsection{Train and Use Meta-learner}
For each ensemble+HP we use the train set with the respective target column $y(ensemble+HP)$. Because class $1$ (select) is underrepresented, we balance the dataset using random oversampling. We then fit a classification pipeline for the ensemble+HP, using the Automated Machine Learning framework tpot \cite{olson2016} for algorithm selection and hyperparameter tuning.

If we want to select the ensemble+HPs for a new dataset we extract the meta-features as above and run the meta-learner for each ensemble+HP. The ensemble+HP is selected if its classifier returns 1. In total, the meta-learner selects $n$ ensemble+HPs, where $n \neq K$.

\subsection{Validation - Experiment 2}\label{sec:valmeta}
\subsubsection{Research Questions}
\begin{enumerate}[label=(\roman*)]
\item Does the meta-learner improve the selection process compared to baseline methods?
\item Did we select valuable meta-features for training the meta-learner?
\end{enumerate}
\subsubsection{Methodology}
We extract the dataset for training the meta-learner from the results from Experiment 1, see \ref{sec:metadata}. The training set contains 14690 samples.
We test the meta-learner for various configuration parameters $K$, where we set $K=[1, 2, \dots, 30]$. We test the meta-learner on 1632 datasets, previously unseen to the meta-learner.
For each value of $K$, for each dataset, we extract the rank $r$ of the best selected ensemble+HP method. We then compute the average number of chosen methods across all datasets $N = \text{avg}(n)$ and the average of best rank $R = \text{avg}(r)$ across all datasets. Please be aware that $n \neq K$.
We compare our meta-learner to two baselines: Autorank-based selection, which selects for each dataset the $n$ ensemble+HPs which ranked best in Experiment 1; and random selection, which selects for each dataset $n$ random ensemble+HPs.

\subsection{Results and Discussion - Experiment 2}\label{sec:experiments2}
\subsubsection{Results}
\begin{figure*}[!t]
\begin{minipage}{.5\textwidth}
	\centering
	\includegraphics[width=0.8\columnwidth]{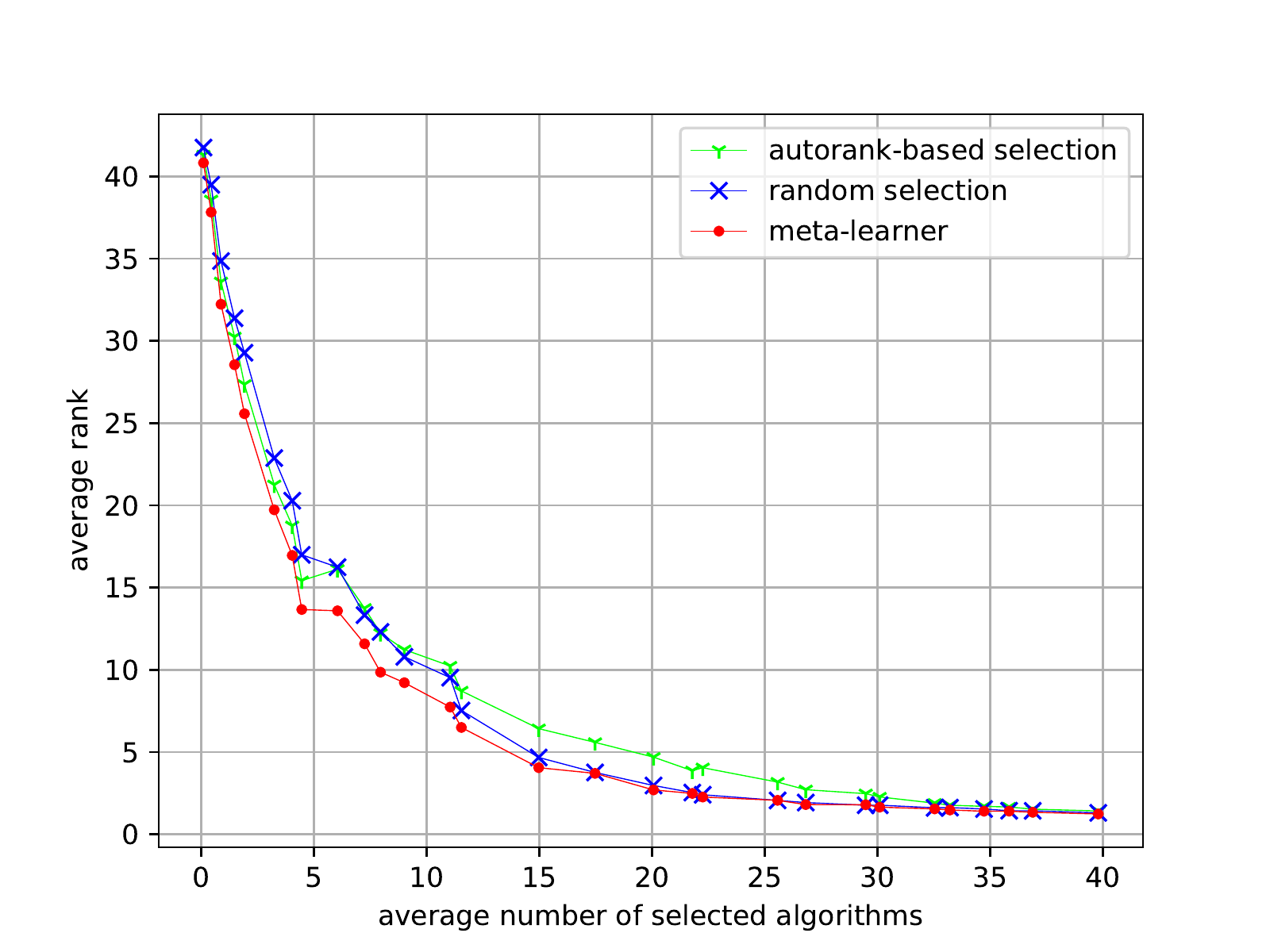}
	
\end{minipage}
\hfill
\begin{minipage}{.5\textwidth}
	\centering
	\includegraphics[width=0.8\columnwidth]{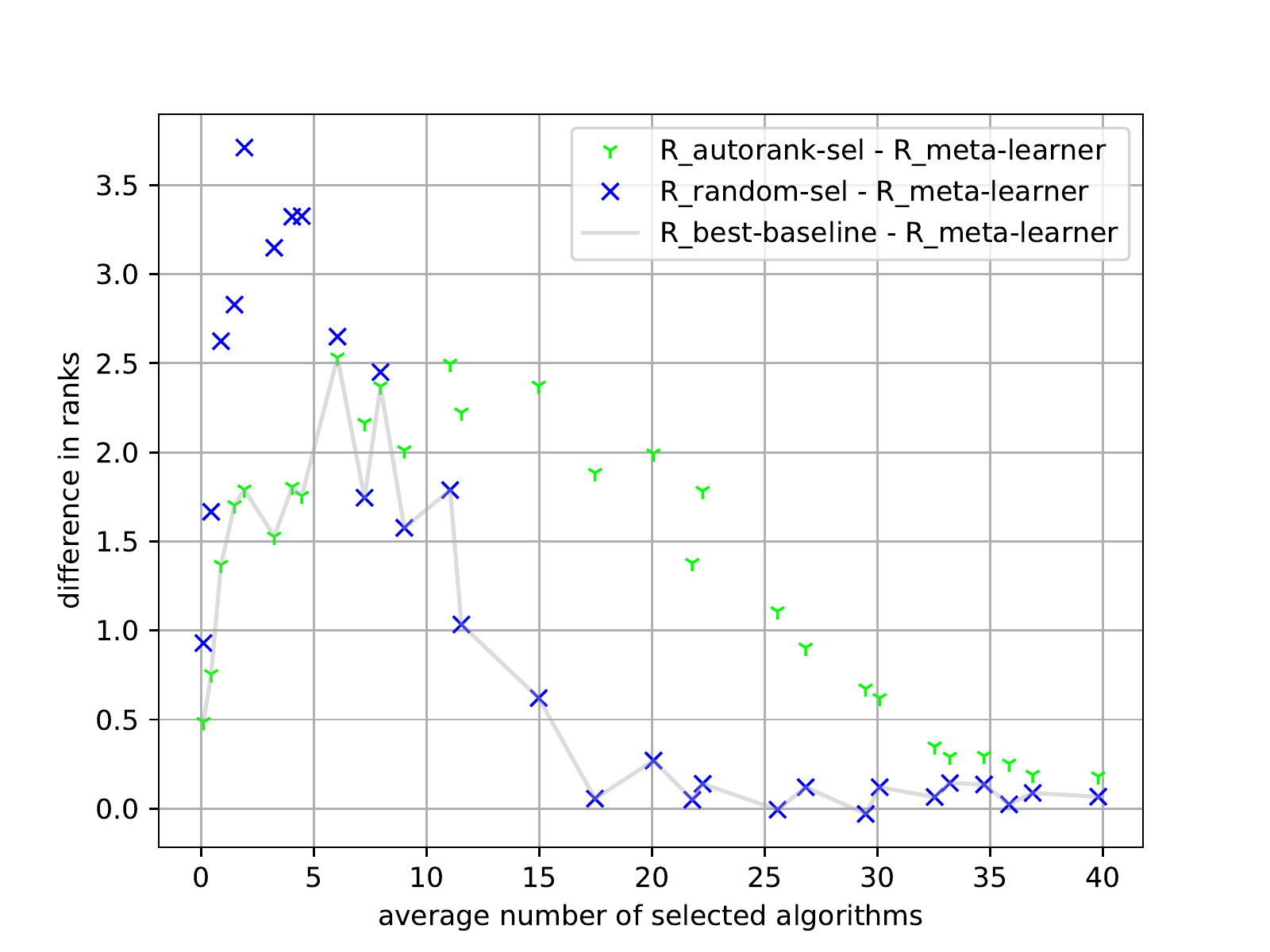}
\end{minipage}
\caption{(a) Average Best Rank $R$ per average number of ensemble+HPs $N$ with different selection strategies and (b) Average Difference in $R$ per $N$ between meta-learner and baselines.}
\label{fig:metaranks}
\end{figure*}

We show a comparison of the meta-learner and the baselines in figure \ref{fig:metaranks}. Figure \ref{fig:metaranks}(a) illustrates the average best rank $R$ across all datasets for the average number of selected ensemble+HPs  $N = \text{avg}(n)$ for a given $K$ (described in section \ref{sec:metadata}) for the meta-learner, the autorank-based selection, and the random selection. Figure \ref{fig:metaranks}(b) shows the difference between the ranks ($R$(baseline) - $R$(meta-learner)) across $N$. The grey line highlights the difference between the lower-ranking baseline and the meta-learner across $N$. We observe that the difference in ranks is mainly positive, and thus that the meta-learners selected ensemble+HPs reach in average lower ranks. The grey line shows maximum values for $4.5\leq N\leq11$. For this interval the best meta-learner-selected ensemble+HPs reach an average rank of ${\sim}13.5$ for $N{\approx}4.5$ and ${\sim}7.5$ for $N{\approx}11$. When selecting less than two ensemble+HPs, we observe $R\text{(meta-learner)}>25$. When selecting more than ten ensemble+HPs, we observe $R\text{(meta-learner)}\leq 8$.

In an additional step, we investigated in feature importance by evaluating the average amount of features used by a random forest meta-learner: We found that all except seven meta-features were used by the random forest\footnote{The figure showing the average feature importance is omitted for space reasons and available on request.}.
%

\subsubsection{Discussion}
In the following, we reply to the research questions in section \ref{sec:valmeta}:

(i) Yes, the meta-learner improves the selection compared to both baseline methods. The results for the meta-learner show lower ranks than both baselines. Thus we recommend using the meta-learner. We still see room for improvement of the meta-learner, especially for the situation of selecting only one ensemble+HP.
Figure \ref{fig:metaranks}(b) shows the highest differences on the grey line for $4.5 \leq N \leq 11$. This means that for this number of selected ensemble+HPs the meta-learner is most beneficial.
In addition, we highlight that the random selection leads to in average lower ranks than the autorank-based selection. 
We see the reason for this in the findings from experiment 1: Horses for courses, the different performance of ensemble+HPs across data-sources, the relatively high winning ranks (i.e. no majorly winning ensembles), and the fact that ensemble+HPs with a low mean rank do not necessarily lead to many lowest ranks. 

(ii) From the feature importance investigation we conclude that our selected meta-features capture the problem satisfactorily. 
In the future, we will remove the meta-features which were not used by the random forest.

\section{Conclusion and Future Work}\label{sec:conclusion}
In this paper, we investigated the performance of ensemble learning methods for time series forecasting, showing results on ${\sim}16000$ time series from various sources. We found that (a) ensemble methods show a benefit in overall forecasting accuracy, with simple ensemble methods leading to in average good results. In addition, we found that (b) it is not clear which ensemble methods to best select with the goal of reaching the best possible forecasting accuracy for automated domain-agnostic time series forecasting, and that, (c) the winning methods (according to average rank) differ for the different data-sources. 
Thus, as a second step, we suggested and evaluated a meta-learning approach. The meta-learner selects for a given dataset, based on its dataset meta-features, a promising subset of ensemble methods plus hyperparameter configurations to run. Results show that the meta-learner outperforms two simple baselines. 
Whereas the results are promising and speak for using a meta-learner, we are positive that the meta-learning results can be improved in future work.
Thus, as future work, we want to investigate in different, improved meta-learning strategies. 
In addition, motivated by the good results of the \textit{superbooster} (Exogenous Variable Postprocessor), we intend to include machine learning base models to even further improve forecasting accuracy on time series with exogenous variables.

\section*{Acknowledgment}
We thank Hamza Saghir for his valuable work on seasonality detection and the inclusion of exogenous Fourier terms in the arima method.

\bibliography{ijcnn2021-refs}
\bibliographystyle{IEEEtran} 

\end{document}